\pdfoutput=1
\def\deanonmode{1}
\documentclass{article}
\PassOptionsToPackage{table}{xcolor}
\newif\ifarxivformat
\newif\ifneuripsformat
\ifdefined\arxivmode
  \arxivformattrue
  \neuripsformatfalse
\else
  \arxivformatfalse
  \ifdefined\iclrmode
    \neuripsformatfalse
  \else
    \neuripsformattrue
  \fi
\fi
\ifarxivformat
  \usepackage[margin=1in]{geometry}
  \usepackage{times}
  \usepackage{natbib}
\else
  \ifneuripsformat
    \ifdefined\deanonmode
      \usepackage[main,preprint]{neurips_2026}
    \else
      \usepackage[main]{neurips_2026}
    \fi
  \else
    \usepackage{template/iclr2026_conference,times}
  \fi
\fi
\usepackage[utf8]{inputenc}
\usepackage[T1]{fontenc}
\usepackage{hyperref}
\ifdefined\deanonmode
  \hypersetup{hidelinks}
\fi
\usepackage{url}
\usepackage{booktabs}
\usepackage{enumitem}
\usepackage{amsfonts}
\usepackage{amsmath}
\usepackage{amssymb}
\IfFileExists{dsfont.sty}{\usepackage{dsfont}}{}

\usepackage{nicefrac}
\usepackage{microtype}
\usepackage[table]{xcolor}
\usepackage{graphicx}
\usepackage{float}
\usepackage{placeins}
\usepackage{subcaption}
\AtBeginEnvironment{table}{\vspace{3pt}}
\AtBeginEnvironment{table*}{\vspace{3pt}}
\usepackage{tikz}

\ifdefined\deanonmode
  \makeatletter
  \renewcommand{\@notice}{}
  \makeatother
\fi
\usepackage{listings}
\definecolor{codekeyword}{RGB}{0,0,255}
\definecolor{codecomment}{RGB}{95,99,104}
\definecolor{codestring}{RGB}{163,21,21}
\definecolor{codefunction}{RGB}{0,90,156}
\definecolor{codeuserfunction}{RGB}{111,66,193}
\definecolor{codesign}{RGB}{35,35,35}
\definecolor{diffred}{RGB}{190,55,55}
\definecolor{diffgreen}{RGB}{28,135,75}
\definecolor{diffredbg}{RGB}{255,225,225}
\definecolor{diffgreenbg}{RGB}{222,247,229}
\definecolor{ourscol}{rgb}{0.86,0.92,0.99}

\lstdefinelanguage{PythonFuncColor}{
  language=Python,
  alsoletter={_},
  keywordstyle=\color{codekeyword},
  commentstyle=\color{codecomment},
  stringstyle=\color{codestring},
  showstringspaces=false,
  literate=*
    {==}{{\color{codesign}== }}{2}
    {<}{{\color{codesign}< }}{1}
    {+=}{{\color{codesign}+= }}{2}
    {self_conditioned_rollout}{{\color{codeuserfunction}self\_conditioned\_rollout}}{24}
    {denoiser}{{\color{codeuserfunction}denoiser}}{8}
    {stopgrad}{{\color{codefunction}stopgrad}}{8}
    {uniform}{{\color{codefunction}uniform}}{7}
    {range}{{\color{codefunction}range}}{5}
    {linspace}{{\color{codefunction}linspace}}{8}
    {update}{{\color{codefunction}update}}{6}
    {maybe_drop}{{\color{codefunction}maybe\_drop}}{10}
    {sampler_step}{{\color{codefunction}sampler\_step}}{12}
    {cross_entropy}{{\color{codefunction}cross\_entropy}}{13}
}
\lstdefinestyle{trainingalgorithm}{
  language=PythonFuncColor,
  basicstyle=\fontsize{7.4pt}{8.2pt}\ttfamily\selectfont,
  columns=fullflexible,
  keepspaces=true,
  showstringspaces=false,
  breaklines=true,
  frame=lines,
  framesep=2mm,
  xleftmargin=1mm,
  xrightmargin=1mm,
  aboveskip=2pt,
  belowskip=2pt
}
\newcommand{\figorplaceholder}[2]{%
  \IfFileExists{#1}{\includegraphics[width=\linewidth]{#1}}{%
    \fbox{\begin{minipage}[c][#2][c]{0.92\linewidth}%
      \centering\itshape\color{gray} [figure pending: \detokenize{#1}]%
    \end{minipage}}}}

\title{Flow Reasoning Models:\\ Turning Flows Into Efficient Recurrent Reasoners}
\author{%
  \textbf{Alec Helbling}\textsuperscript{\textnormal{1,3,4,*}} \quad
  \textbf{Andrey Bryutkin}\textsuperscript{\textnormal{2,3,4,*}} \quad
  \textbf{Mauro Martino}\textsuperscript{\textnormal{3,4}} \\
  \vspace{0.2em} \\
  \textbf{Duen Horng (Polo) Chau}\textsuperscript{\textnormal{1}} \quad
  \textbf{Nima Dehmamy}\textsuperscript{\textnormal{3,4}} \quad
  \textbf{Hendrik Strobelt}\textsuperscript{\textnormal{3,4}} \\
  \vspace{0.5em} \\
  \textsuperscript{1}Georgia Tech \quad
  \textsuperscript{2}MIT \quad
  \textsuperscript{3}MIT-IBM Computing Research Lab \quad
  \textsuperscript{4}IBM Research
}
\ifarxivformat\date{}\fi

\begin{document}

\maketitle
\begingroup
\renewcommand{\thefootnote}{}
\footnotetext{\textsuperscript{*}Co-first author.\quad Correspondence to Alec Helbling: \texttt{alechelbling@gatech.edu}}
\addtocounter{footnote}{-1}
\endgroup
\begingroup
\renewcommand{\thefootnote}{}
\footnotetext{See code at: \url{https://github.com/helblazer811/Flow-Reasoning-Models}}
\addtocounter{footnote}{-1}
\endgroup
\ifdefined\deanonmode\enlargethispage{2\baselineskip}\fi

\begin{abstract}
Structured reasoning requires making and revising interdependent decisions to reach a globally
consistent solution. Existing architectures struggle with this: autoregressive models commit
sequentially and cannot revise earlier decisions, while masked diffusion models often require careful decoding schemes
to coordinate interdependent predictions. We introduce Flow Reasoning Models (FRMs), a novel
framework for structured reasoning that adapts continuous flows over discrete structured outputs with a simple recurrent refinement
mechanism. By self-conditioning a flow model on its own past outputs, we turn one-shot denoising
into iterative solution refinement. This lets FRMs make and revise decisions in parallel,
efficiently coordinating interdependent choices across solutions. Yet conventional
self-conditioning becomes unreliable at greater recurrent depth due to exposure bias between
one-step training predictions and recursively generated inference states. We address this
mismatch with Fixed-Point Forcing (FPF), which trains FRMs on states produced by their own
inference dynamics while preserving the standard flow-matching objective. FRMs achieve solve rates
of $99.5\%$, $100.0\%$, and $99.9\%$ on Sudoku-Extreme, Zebra, and Maze-Unique, respectively. On
Sudoku-Extreme, FRMs achieve higher peak accuracy than the evaluated masked-diffusion and
specialized reasoning baselines while remaining highly compute-efficient, matching the next-best
method's $98.7\%$ peak solve rate with $44\times$ fewer inference FLOPs.
\end{abstract}
\section{Introduction}
\label{sec:intro}

\begin{figure*}[t]
  \centering
  \includegraphics[width=\textwidth]{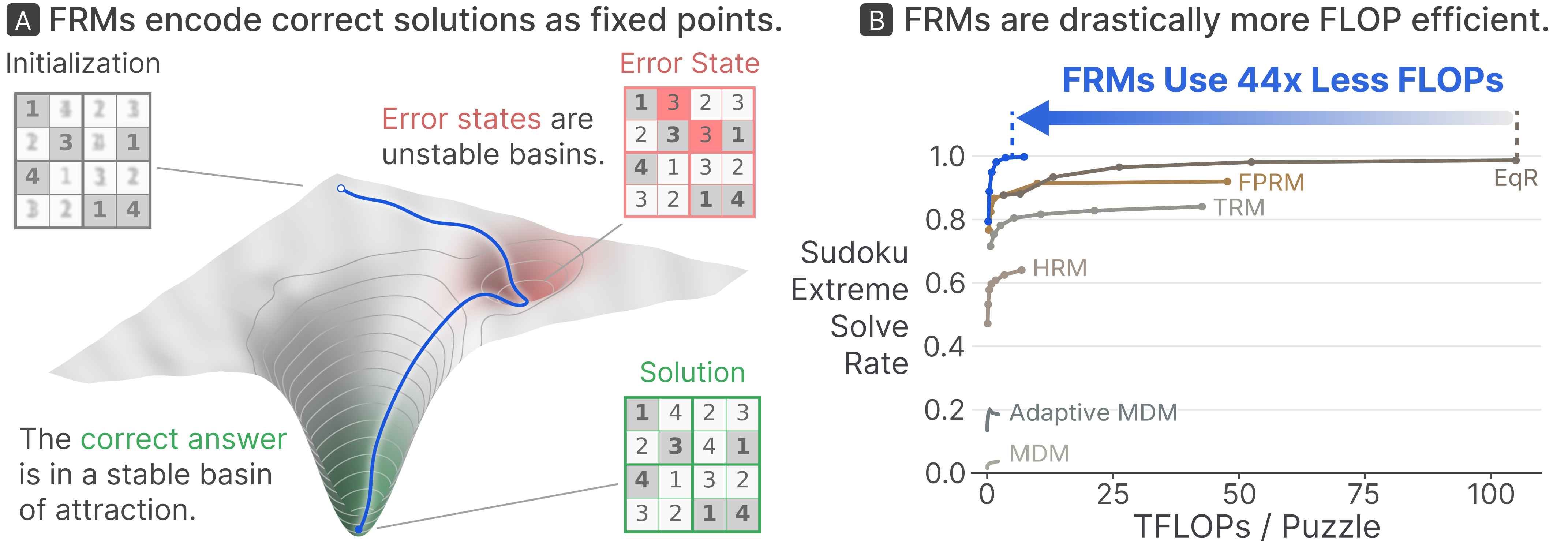}
  \caption{\textbf{Flow Reasoning Models turn stable recurrent dynamics into efficient
  reasoning.} Left: a conceptual view of the recurrent solution landscape. Error states occupy
  unstable basins, while the correct solution forms a stable attractor that recurrent sampling
  converges toward. Right: solve rate versus inference FLOPs on
  Sudoku-Extreme. At the highlighted matched solve rate, FRM uses $44\times$ fewer FLOPs than the
  comparison method. The landscape is an explanatory schematic rather than a literal
  three-dimensional embedding of model states.}
  \label{fig:crown-jewel}
\end{figure*}

A central challenge in structured reasoning is coordinating many mutually constraining decisions
into a single consistent solution. Autoregressive models often struggle with such tasks because
they commit to tokens in a fixed order, without the ability to later revise them as their global
consequences become clear. Masked diffusion models~\citep{sahoo2024simpleeffectivemaskeddiffusion} offer a flexible
generation order and predict multiple tokens in parallel, but tokens updated within the same step
are conditionally independent given the current state and cannot condition on one another's
realized values, making aggressive parallel updates prone to
inconsistencies~\citep{kim2025trainworstplanbest}. Tightly constrained problems therefore require
many conservative updates, and correcting earlier errors requires additional mechanisms such as
remasking~\citep{wang2026remaskingdiscretediffusionmodels}.

Continuous flow models over discrete data offer a promising alternative, modeling all tokens and their
interdependencies simultaneously. Recent work has demonstrated that continuous flow models can
effectively perform unconditional generation of natural language~\citep{lee2026flowmaplanguagemodels,hu2026elfembeddedlanguageflows},
but their ability to perform structured reasoning remains largely unexplored. Indeed, in our own experiments, a naive application of such continuous flows
to structured reasoning tasks like Sudoku solves only approximately $13\%$ on the
Sudoku-Extreme benchmark~\citep{wang2025hierarchicalreasoningmodel}.

We introduce Flow Reasoning Models (FRMs), which turn continuous flows over discrete outputs into recurrent reasoning
models that iteratively refine candidate solutions. Self-conditioning has previously been used to
improve unconditional generation~\citep{chen2023analogbitsgeneratingdiscrete,hu2026elfembeddedlanguageflows}.
We observe that it can be particularly useful for decomposing complex reasoning into successive
refinement steps: FRMs feed each prediction back into the flow as the state to be refined at the
next step. This yields an interpretable recurrent state
that can be directly decoded into the model's current candidate solution. It also provides a
training-efficient form of recurrence that
requires no backpropagation through time and retains the canonical flow-matching objective.
Adding self-conditioning alone significantly improves performance, raising the solve rate on
Sudoku-Extreme from roughly $13\%$ to $33\%$, but still leaving most problems unsolved.

This limitation becomes clearer through a dynamical-systems interpretation of FRMs: recurrent refinement performs
learned fixed-point iteration over candidate solutions. Under this view, correct solutions should
form stable, error-correcting attractors, while incorrect states should remain transient.
Conventional self-conditioning fails to produce these dynamics on challenging problems, instead
converging to confidently incorrect fixed points. We trace this failure to exposure bias between
the one-step conditioning states used during training and the recursively generated states
encountered during
inference~\citep{bengio2015scheduledsampling,huang2025selfforcingbridgingtraintest}. We introduce
Fixed-Point Forcing (FPF), which trains on rollout-derived conditioning states so that the model
learns to correct states produced by its own dynamics. Our results support this attractor picture:
under FPF, recurrent refinement moves predictions toward the ground truth, correct solutions
converge while incorrect states remain dynamically active, and convergence becomes strongly
predictive of correctness.

\noindent Our core contributions are as follows:
\begin{enumerate}[leftmargin=1.5em]
  \item \textbf{Flow Reasoning Models (FRMs), a novel generative modeling framework that turns
        continuous flows over discrete structured outputs into recurrent models for reasoning.} Their
        predictions form an interpretable, directly decodable reasoning state while the models
        retain the canonical flow-matching objective and require no backpropagation through time.
  \item \textbf{Fixed-Point Forcing (FPF), a novel training method for recurrent self-conditioned
        flow models.} We discover that exposure bias can cause these models to converge to spurious
        fixed points. FPF mitigates this mismatch by training on rollout-derived conditioning
        states, making additional recurrent depth productive and convergence predictive of
        correctness.
  \item \textbf{State-of-the-art accuracy and efficiency across structured reasoning tasks.}
        Across Sudoku-Extreme, Maze, and Zebra, FRMs achieve state-of-the-art solve rates and compare
        favorably with masked diffusion models and specialized small reasoning models in accuracy
        and inference efficiency. On Sudoku-Extreme, FRMs reach a peak solve rate of $99.5\%$ and
        match EqR's peak $98.7\%$ solve rate~\citep{huang2026equilibriumreasonerslearningattractors}
        with $44\times$ fewer inference FLOPs (Figure~\ref{fig:crown-jewel}).
\end{enumerate}

\begin{figure}[t!]
  \centering
  \includegraphics[width=\textwidth]{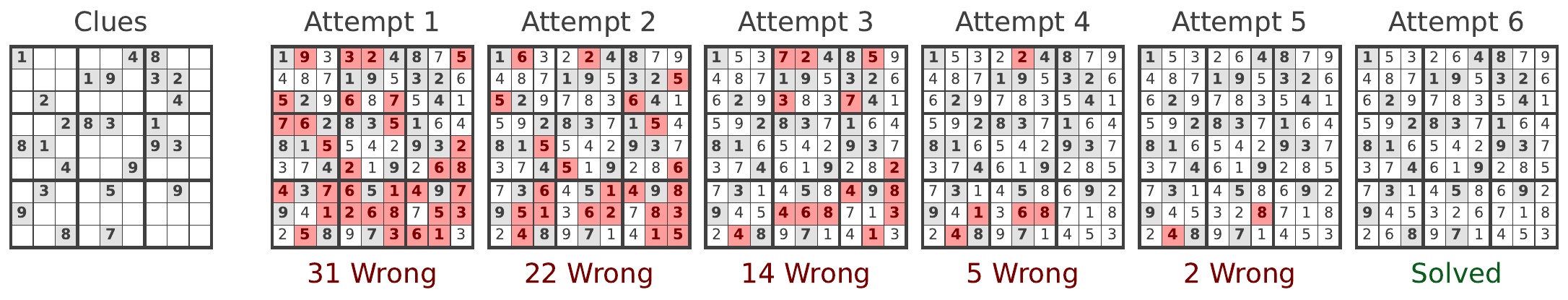}
  \caption{\textbf{Recurrent refinement progressively solves a Sudoku-Extreme instance.}
  Starting from the fixed clues (gray), we decode the FRM's prediction after each recurrent
  attempt, with incorrect entries shown in red. The rollout illustrates how recurrent computation
  revises globally inconsistent early predictions rather than committing to them permanently.}
  \label{fig:sudoku-decoded-refinement}
\end{figure}

\section{Flow Reasoning Models}
\label{sec:method}

\subsection{Discrete Flows for Conditional Reasoning}

We formulate reasoning as conditional generation over data pairs $(c,y)\sim p_{\mathrm{data}}$.
Here $c$ denotes the observed problem specification, for example, the given clues in Sudoku or
the maze layout, and $y=(y_1,\ldots,y_L)$ denotes its discrete solution over vocabulary
$\mathcal V$. Only the solution is noised and generated: $c$ is supplied as a separate conditioning
input and held fixed throughout training and inference. Following work that models discrete data through
continuous or simplex-valued representations
\citep{austin2021structured,gat2024discreteflowmatching,chen2023analogbitsgeneratingdiscrete,
stark2024dirichletflowmatching}, we encode $y$ as a one-hot endpoint
$x_1\in\{0,1\}^{L\times|\mathcal V|}$ and decode predictions by positionwise argmax.

Flow matching connects Gaussian noise $\varepsilon\sim\mathcal N(0,I)$ to this endpoint along a
probability path~\citep{lipman2023flowmatching,albergo2025stochasticinterpolants}. We use the linear
interpolant
\begin{equation}
  x_t=(1-t)\varepsilon+t x_1, \qquad t\in[0,1].
  \label{eq:frm-path}
\end{equation}
We adopt the categorical clean-prediction parameterization of Flow
Language Models~\citep{lee2026flowmaplanguagemodels}: a denoiser
$D_\theta^t(x_t\mid c)$ predicts a categorical distribution over the clean token at each position.
The corresponding clean endpoint prediction determines the probability-flow velocity:
\begin{equation}
  v_\theta^t(x_t\mid c)=\bigl(D_\theta^t(x_t\mid c)-x_t\bigr)/(1-t).
  \label{eq:frm-velocity}
\end{equation}
The same output is therefore both a directly decodable candidate solution and the quantity used to
advance the flow. We train it with tokenwise cross-entropy,
\begin{equation}
  \mathcal L_{\mathrm{CE}}(\theta)=
  \mathbb E_{t,c,y,\varepsilon}\!\left[-\sum_{i=1}^{L}
  \log D_\theta^t(x_t\mid c)_{i,y_i}\right].
  \label{eq:frm-ce}
\end{equation}
This standard discrete-flow model is our FLM baseline. On its own it leaves a substantial gap on
structured reasoning; FRMs add a recurrent axis while retaining this path and objective.

\subsection{Self-Conditioning as Recurrent Reasoning}

\paragraph{Self-conditioning.}
Self-conditioning augments a denoiser with its own previous output as an additional input
\citep{chen2023analogbitsgeneratingdiscrete}. We write the resulting model as
$D_\theta^t(x_t\mid c,s)$, where $s$ carries a previous clean-solution prediction and
$s=\varnothing$ denotes a zero-valued null carry. Conventional training uses two passes: a
null-carry pass produces the detached prediction
$\widetilde{s}=\operatorname{stopgrad}[D_\theta^t(x_t\mid c,\varnothing)]$, and the supervised pass
receives either $\widetilde{s}$ or the null carry. The probability path and endpoint loss in
Eq.~\eqref{eq:frm-ce} remain unchanged, and gradients do not pass through $\widetilde{s}$.

\paragraph{Self-conditioning creates recurrence.}
At inference, FRMs repeatedly feed the latest clean prediction back through $s$. This feedback lets
later evaluations preserve useful decisions, revise inconsistent ones, and correct earlier
mistakes. It also creates a discrete reasoning depth $k$, separate from continuous flow time $t$.
Holding $(x_t,t)$ fixed, recurrent refinement is
\begin{equation}
  s_t^{(0)}=\varnothing,
  \qquad
  s_t^{(k+1)}=D_\theta^t(x_t\mid c,s_t^{(k)}),
  \label{eq:frm-recurrence}
\end{equation}
so each carry directly estimates the same clean solution. This gives every update direct
supervision, allowing detached training without backpropagation through time. A sampler can
alternate between advancing the flow state $x_t$ and applying additional recurrent updates at the
current flow time. Although self-conditioning substantially improves refinement, its gains saturate
with depth on Sudoku-Extreme
(Fig.~\ref{fig:fixed-point-stability-triptych}b), motivating the fixed-point analysis and training
method developed next.
\begin{figure*}[!t]
  \centering
  \begin{subfigure}[t]{0.485\textwidth}
    \centering
    \includegraphics[width=\linewidth]{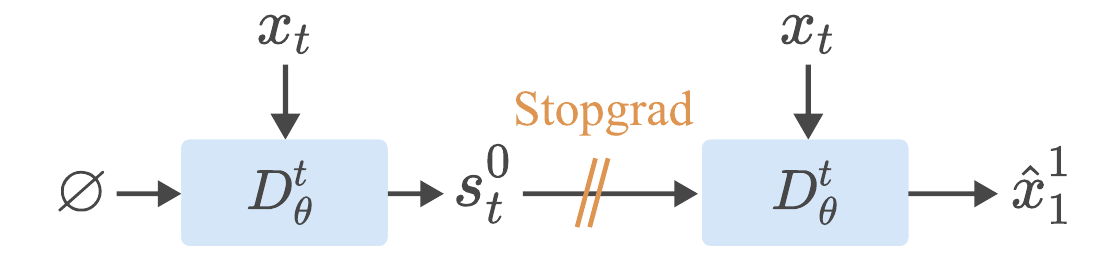}
    \caption{Conventional self-conditioning}
  \end{subfigure}
  \hfill
  \begin{subfigure}[t]{0.485\textwidth}
    \centering
    \includegraphics[width=\linewidth]{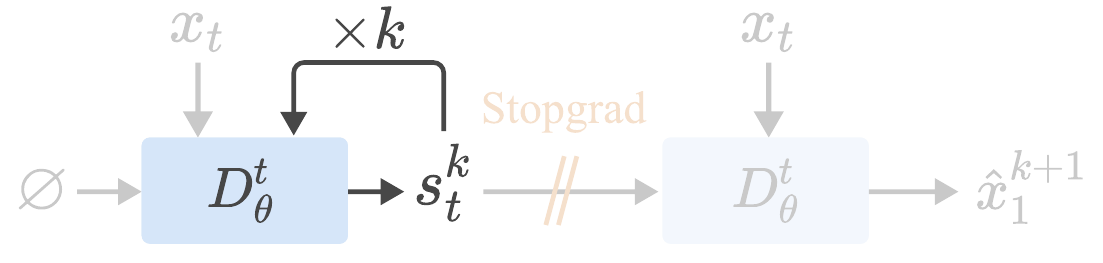}
    \caption{Fixed-Point Forcing}
  \end{subfigure}
\caption{\textbf{Fixed-Point Forcing replaces one-pass carries with rollout-derived carries.}
Conventional self-conditioning forms a detached carry with one denoiser pass at the supervised
interpolant. Fixed-Point Forcing instead forms the detached carry through repeated application of
the denoiser, exposing the loss-bearing update to a state produced by the model's recurrent
inference dynamics.}
\label{fig:self-conditioning-fpf-diagrams}
\end{figure*}

\section{Fixed-Point Forcing Promotes Correct Solutions as Stable Attractors}
\label{sec:fixed-point-dynamics}

The recurrence in Eq.~\eqref{eq:frm-recurrence} defines a dynamical system over candidate
solutions. Fixing $(x_t,c,t)$, repeated application of $D_\theta^t$ performs fixed-point iteration
over the self-conditioning state. During an FPF rollout, however, the flow state, time, and carry
evolve jointly, so this fixed-point iteration is a held-state view rather than a literal description
of how the FPF carry is constructed. Flow time moves through a family of denoising problems, while
reasoning depth iterates the corresponding denoiser toward a stable prediction. For a problem $c$
with goal $y^\star$, the desired held-state behavior is
\begin{equation}
  s_t^{(k)}\longrightarrow s_t^\star,
  \qquad
  D_\theta^t(x_t\mid c,s_t^\star)=s_t^\star,
  \qquad
  \operatorname{decode}(s_t^\star)=y^\star.
  \label{eq:frm-solution-fixed-point}
\end{equation}
A useful solution should be attracting: imperfect nearby states should move toward it under
additional reasoning depth. Stable states that decode to incorrect solutions are instead
\emph{spurious fixed points}. This perspective supplies a common language for the rest of the
paper: training determines the recurrent state distribution, inference probes the resulting
closed-loop dynamics, and useful test-time computation requires those dynamics to correct rather
than stabilize errors.

\subsection{Exposure Bias in the Recurrent State}

Conventional self-conditioning trains with a detached one-pass carry produced from the
ground-truth-derived interpolant $x_t=(1-t)\varepsilon+t y$, with
$\widetilde{s}=\operatorname{stopgrad}[D_\theta^t(x_t\mid c,\varnothing)]$
(Fig.~\ref{fig:self-conditioning-fpf-training-algorithms}, left). Recurrent inference instead
feeds predictions back repeatedly, so the carry at depth $k$ is generated by the model's own
closed-loop dynamics rather than by one pass from a ground-truth-derived state. Training and
inference therefore induce different carry distributions,
\begin{equation}
  p_{\mathrm{train}}(s\mid x_t,c,t)
  \neq
  p_{\mathrm{infer}}^{(k)}(s\mid \widehat{x}_t,c,t),
  \label{eq:frm-exposure-bias}
\end{equation}
and the mismatch grows with depth.

This mismatch has two coupled effects. First, the denoiser is not calibrated for its own recurrent
states: an incorrect but overconfident prediction can be fed back as reliable evidence, amplify its
error, and settle at a spurious fixed point. Second, one-pass training rarely presents the subtle,
low-residual errors that remain after several refinement steps, so the model is not trained to
correct precisely the states encountered near convergence. On challenging problems, recurrence can
therefore stabilize and sharpen an incorrect solution, increasing gold-target cross-entropy with depth
(Fig.~\ref{fig:fixed-point-stability-triptych}).

\subsection{Fixed-Point Forcing}

Fixed-Point Forcing (FPF) replaces the one-pass carry used in conventional self-conditioning with a
detached carry produced by the model's own multi-step inference dynamics. It thereby trains each
recurrent update on the kinds of states produced by earlier updates, directly targeting the
train--inference mismatch in Eq.~\eqref{eq:frm-exposure-bias}.

\begin{figure}[H]
\centering
\captionsetup{aboveskip=3pt}
\setlength{\fboxsep}{0pt}
\noindent\makebox[\linewidth][c]{Conventional Self-Conditioning $\longrightarrow$ Fixed-Point Forcing}\par
\vspace{0.45em}
\begingroup
\raggedright
\noindent\begin{tikzpicture}[every node/.style={inner sep=0pt,outer sep=0pt}]
  \node[anchor=west] (old) at (0,0) {\colorbox{diffredbg}{\parbox[c][3.45em][c]{0.40\textwidth}{%
    \fontsize{8.4pt}{10.4pt}\selectfont\ttfamily
    \hspace*{0.6em}\textcolor{diffred}{- }pred = \textcolor{codeuserfunction}{denoiser}(\\
    \hspace*{0.6em}\textcolor{diffred}{- }\hspace*{1em}x\_t, t, cond=clues, carry=\textcolor{codekeyword}{None}\\
    \hspace*{0.6em}\textcolor{diffred}{- })}}};
  \node[anchor=south west] at ([yshift=0.2em]old.north west) {\begin{minipage}{0.54\textwidth}
\begin{lstlisting}[style=trainingalgorithm,frame=none,numbers=none,xleftmargin=0pt,aboveskip=0pt,belowskip=0pt,basicstyle=\normalfont\fontsize{8.4pt}{10.4pt}\ttfamily\selectfont]
t = uniform(0, 1)
x_t = (1 - t) * noise + t * target
\end{lstlisting}
\end{minipage}};
  \node[anchor=west] (new) at (0.51\textwidth,0) {\colorbox{diffgreenbg}{\parbox[c]{0.49\textwidth}{%
    \vspace{0.35em}%
    \fontsize{8.4pt}{10.4pt}\selectfont\ttfamily
    \hspace*{0.6em}\textcolor{diffgreen}{+ }\textcolor{codecomment}{\# Construct a rollout-derived carry.}\\
    \hspace*{0.6em}\textcolor{diffgreen}{+ }t\_start = \textcolor{codefunction}{uniform}(0, t)\\
    \hspace*{0.6em}\textcolor{diffgreen}{+ }x\_start = (1 - t\_start) * noise\\
    \hspace*{0.6em}\textcolor{diffgreen}{+ }\hspace*{4.9em}+ t\_start * target\\
    \hspace*{0.6em}\textcolor{diffgreen}{+ }pred = \textcolor{codeuserfunction}{self\_conditioned\_rollout}(\\
    \hspace*{0.6em}\textcolor{diffgreen}{+ }\hspace*{1em}x\_start, t\_start, t, cond=clues\\
    \hspace*{0.6em}\textcolor{diffgreen}{+ })\vspace{0.35em}}}};
  \shade[left color=diffredbg,right color=diffgreenbg]
    (old.north east) -- (new.north west) -- (new.south west) -- (old.south east) -- cycle;
  \node[anchor=north west] at ([yshift=-0.25em]old.south west) {\begin{minipage}{0.54\textwidth}
\begin{lstlisting}[style=trainingalgorithm,frame=none,numbers=none,xleftmargin=0pt,aboveskip=0pt,belowskip=0pt,basicstyle=\normalfont\fontsize{8.4pt}{10.4pt}\ttfamily\selectfont]
carry = stopgrad(pred)
refined_pred = denoiser(
    x_t, t, cond=clues, carry=carry
)
update(cross_entropy(refined_pred, target))
\end{lstlisting}
\end{minipage}};
\end{tikzpicture}\par
\endgroup
\caption{\textbf{Conventional self-conditioning and Fixed-Point Forcing training.}
Both procedures supervise the same canonical interpolant. Conventional self-conditioning obtains
its carry from one pass at that interpolant, whereas FPF obtains a detached carry from a recursive
rollout whose start time is sampled uniformly before the supervision time and whose endpoint is the
supervision time itself. The \texttt{self\_conditioned\_rollout} helper denotes the standard
self-conditioned integration loop used at inference and for constructing the FPF carry.}
\label{fig:self-conditioning-fpf-training-algorithms}
\end{figure}

The conceptual difference is illustrated in
Fig.~\ref{fig:self-conditioning-fpf-diagrams}, and the corresponding training-code change is shown
in Fig.~\ref{fig:self-conditioning-fpf-training-algorithms}. We construct the FPF conditioning state
by sampling a supervision time $t$ and a rollout start
$t_{\mathrm{start}}\sim\mathcal{U}(0,t)$, then run the inference-time self-conditioned integrator
from $t_{\mathrm{start}}$ to $t$. Its final prediction $s_{\mathrm{FPF}}$ is detached and supplied to
the loss-bearing prediction
$D_\theta^t(x_t\mid c,\operatorname{stopgrad}(s_{\mathrm{FPF}}))$; gradients do not pass through the
rollout.

Crucially, this construction preserves the canonical flow-matching path exactly. As in conventional
self-conditioning, the loss-bearing input remains $x_t=(1-t)\varepsilon+t y$, the target remains
$y$, and the endpoint cross-entropy is unchanged; the rollout prediction enters only through the
self-conditioning channel. FPF therefore changes the distribution of what the denoiser conditions
on, not the state at which the flow objective is supervised. Training on rollout-derived carries
reduces the carry-side exposure bias and the resulting overconfidence and miscalibration,
while deeper rollout states expose the model to the smaller, subtler residual errors that remain
near convergence and train the local corrections needed around a fixed point.

\section{Experiments}
\label{sec:experiments}

\begingroup
\makeatletter
\renewcommand{\paragraph}{%
  \@startsection{paragraph}{4}{\z@}%
                {0.9ex \@plus 0.3ex \@minus 0.2ex}%
                {-1em}%
                {\normalsize\bf}%
}
\makeatother
We evaluate Flow Reasoning Models (FRMs) on Sudoku-Extreme, Zebra, and Maze-Unique, three
structured-prediction benchmarks with different constraints and output structures. Our primary
metric is exact solve rate. We compare inference efficiency using total FLOPs per instance and
report the number of function evaluations (NFE) to describe how each method allocates iterative
computation. Full protocols, baseline provenance, compute accounting, sweeps, and dynamics
diagnostics appear in Appendices~\ref{app:datasets}--\ref{app:mechanism-support}.

\newcommand{\headlineComparisonTable}{%
  \centering
  \small
  \setlength{\tabcolsep}{3pt}
  \begin{tabular}{@{}p{5.1cm}*{6}{>{\centering\arraybackslash}p{1.22cm}}@{}}
    \toprule
    & \multicolumn{2}{c}{\textbf{Sudoku-Extreme}}
    & \multicolumn{2}{c}{\textbf{Zebra}}
    & \multicolumn{2}{c}{\textbf{Maze-Unique}} \\
    \cmidrule(lr){2-3}\cmidrule(lr){4-5}\cmidrule(lr){6-7}
    \textbf{Method}
    & \textbf{Acc. (\%)}
    & \textbf{Size}
    & \textbf{Acc. (\%)}
    & \textbf{Size}
    & \textbf{Acc. (\%)}
    & \textbf{Size} \\
    \midrule
    \multicolumn{7}{@{}l}{\textit{Diffusion language models and flows}} \\
    \addlinespace[2pt]
    FMLM \citep{lee2026flowmaplanguagemodels}
                                        & $1.0$ & $7$M & $74.2$ & $25$M & $87.5$ & $8.1$M \\
    MDLM \citep{sahoo2024simpleeffectivemaskeddiffusion}
                                        & $3.9$ & $30$M & $76.9^{\dagger}$ & $19$M & $89.0$ & $8$M \\
    ReMDM \citep{wang2025remaskingdiscretediffusionmodels}
                                        & $5.5$ & $30$M & -- & -- & $93.5$ & $8$M \\
    FLM \citep{lee2026flowmaplanguagemodels}
                                        & $13.9$ & $7$M & $67.9$ & $25$M & $93.3$ & $8$M \\
    Adaptive MDLM \citep{kim2025trainworstplanbest}
                                        & $19.1$ & $30$M & $98.3^{\dagger}$ & $19$M & $100.0$ & $8$M \\
    \midrule
    \multicolumn{7}{@{}l}{\textit{Specialized recurrent reasoners}} \\
    \addlinespace[2pt]
    HRM \citep{wang2025hierarchicalreasoningmodel}
                                        & $64.1$ & $27.3$M & -- & -- & $0.3$ & $27.3$M \\
    TRM \citep{jolicoeurmartineau2025morerecursivereasoningtiny}
                                        & $84.1$ & $17.6$M & -- & -- & $77.9$ & $0.264$M \\
    FPRM \citep{movahedi2026fixedpointreasoners}
                                        & $94.2$ & $7$M & -- & -- & $100.0$ & $7$M \\
    EqR \citep{huang2026equilibriumreasonerslearningattractors}
                                        & $98.7$ & $5.03$M & -- & -- & $93.0$ & $2.6$M \\
    \midrule
    \multicolumn{7}{@{}c@{}}{%
      \begingroup\setlength{\fboxsep}{0pt}%
      \colorbox{ourscol}{%
        \begin{tabular}{@{}p{5.1cm}*{6}{>{\centering\arraybackslash}p{1.22cm}}@{}}
          \textbf{Flow Reasoning Model (ours)}
          & $99.5$ & $7$M & $100.0$ & $25$M & $99.9$ & $8$M
        \end{tabular}}%
      \endgroup} \\
    \bottomrule
  \end{tabular}
  \captionof{table}{\textbf{Peak exact-solve accuracy and model size across structured reasoning tasks.}
  Parameter counts correspond to the evaluated configuration and are rounded to the nearest
  reported model-size convention. Dashes indicate an unreported result or unavailable
  configuration-matched count. $^{\dagger}$Published Zebra result not reproduced in our evaluation
  pipeline; see Appendix~\ref{app:baseline-accounting}.}
  \label{tab:best-performance-compute}
  \label{tab:recurrent-reasoner-performance-compute}
  \vspace{-0.5em}
}
\begin{figure*}[t!]
  \centering
  \includegraphics[width=\linewidth]{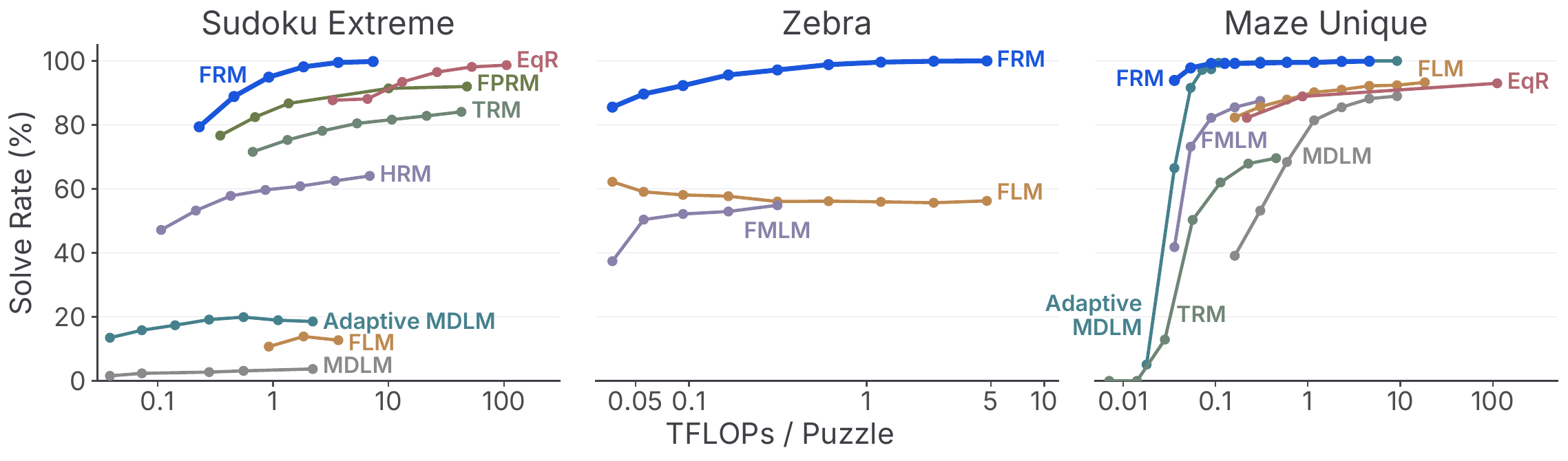}
  \caption{\textbf{Accuracy--compute frontiers across structured-reasoning tasks.}
  Solve rate is plotted against total inference FLOPs per instance for Sudoku-Extreme, Zebra, and
  Maze-Unique. The shared solve-rate axis and legend are shown on the left panel only. Each panel
  compares FRM operating points with the applicable flow baselines under a common FLOP-counting
  convention. We show only method--task pairs with a valid, configuration-matched operating curve;
  unavailable combinations are omitted rather than extrapolated.}
  \label{fig:cross-task-flop-frontiers}
  \vspace{0.6em}
  \headlineComparisonTable
\end{figure*}

\subsection{Accuracy and Efficiency Across Structured Reasoning Tasks}
\label{sec:headline-results}

\paragraph{Flow Reasoning Models achieve near-perfect exact solving across distinct reasoning
domains.}
FRMs reach peak solve rates of $99.5\%$ on Sudoku-Extreme, $100.0\%$ on Zebra, and $99.9\%$ on
Maze-Unique (Table~\ref{tab:best-performance-compute}). These benchmarks require
constraint satisfaction, relational deduction, and path finding, respectively, and differ
substantially in sequence length, vocabulary, and output structure. Near-perfect performance across
all three therefore shows that recurrent flow refinement is not tied to a particular task
representation or constraint structure.

\paragraph{FRMs establish a stronger accuracy--compute frontier than diffusion language models and
specialized recurrent reasoners.}
FRM leads the measured accuracy--compute frontier on all three tasks
(Fig.~\ref{fig:cross-task-flop-frontiers}). On Sudoku-Extreme, it attains the highest solve rate and
reaches EqR's $98.7\%$ peak with an estimated $44\times$ fewer inference FLOPs. FRM likewise
dominates the measured Maze-Unique frontier, although its $99.9\%$ peak is slightly below the best
observed $100.0\%$, and reaches $100.0\%$ on Zebra, exceeding both reported peaks and the measured
baselines. Zebra contains fewer curves because we could not reproduce several published MDM models
reliably enough for FLOP profiling; their daggered, table-only accuracies and full provenance appear
in Table~\ref{tab:best-performance-compute} and
Appendix~\ref{app:baseline-accounting}.

\subsection{Self-Conditioning Enables Recurrent Refinement}
\label{sec:component-results}

\paragraph{Vanilla discrete flows achieve limited peak performance and do not scale with inference
compute.}
Across all three tasks, the vanilla FLM baselines plateau well below the full FRM
(Table~\ref{tab:component-ablation}), and allocating more forward passes leaves their solve-rate
curves essentially flat. Additional integration more accurately traces the learned probability
path, but provides no mechanism for repeatedly revising the current solution.

\paragraph{Self-conditioning enables recurrent refinement.}
Adding self-conditioning improves solve rate over a plain discrete flow on Sudoku-Extreme
($13.1\% \rightarrow 32.6\%$) and Maze-Unique ($77.6\% \rightarrow 96.2\%$;
Table~\ref{tab:component-ablation}). On Zebra, however, self-conditioning lowers the three-seed
mean from $62.3\%$ to $36.9\%$, and both the base and self-conditioned flows exhibit high seed
variance. Feeding each prediction back to
the model creates a recurrent computation that can preserve correct assignments and revise earlier
mistakes, as illustrated by the decoded trajectory in Fig.~\ref{fig:sudoku-decoded-refinement}. On
an easier Sudoku variant introduced by \citet{shah2024causallanguagemodeling} (Sudoku-Shah),
self-conditioning alone is already sufficient to raise solve rate from approximately $30\%$ to
$99\%$, showing that recurrent refinement can saturate less demanding constraint problems without
Fixed-Point Forcing.

\paragraph{Conventional self-conditioning remains unreliable at depth.}
Despite these gains, its Sudoku-Extreme solve rate plateaus near one-third of puzzles as recurrent
depth increases (Fig.~\ref{fig:fixed-point-stability-triptych}b), motivating a training procedure
that makes the recurrent computation reliable over long rollouts.

\subsection{Fixed-Point Forcing Creates Healthy Reasoning Dynamics}
\label{sec:mechanistic-analysis}

\paragraph{Exposure bias produces confidently incorrect fixed points.}
With conventional self-conditioning, solve rate plateaus as recurrent depth increases while
gold-target cross-entropy grows even as the adjacent-state symmetric-KL residual shrinks
(Fig.~\ref{fig:fixed-point-stability-triptych}a,b). The model is therefore not simply failing to
converge: it is becoming increasingly confident in self-consistent but incorrect predictions.
Correspondingly, residual magnitude is uninformative about correctness (AUROC $0.50$;
Fig.~\ref{fig:fixed-point-stability-triptych}c).

\paragraph{Fixed-Point Forcing mitigates exposure bias, enabling recurrent test-time scaling.}
Replacing the one-pass training carry with a rollout-derived carry raises the three-seed mean solve
rate from $32.6\%$ to $99.2\%$ on Sudoku-Extreme and from $96.2\%$ to $98.0\%$ on Maze-Unique.
On Zebra, the three FPF seeds average $99.9\%$ (Table~\ref{tab:component-ablation}); the best
evaluated
operating points reach $99.5\%$ on Sudoku-Extreme and $99.9\%$ on Maze-Unique
(Table~\ref{tab:best-performance-compute}). More importantly, conventional
self-conditioning plateaus with recurrent depth, whereas FPF continues converting additional model
evaluations into corrected solutions (Fig.~\ref{fig:fixed-point-stability-triptych}b). Because FPF
changes the training distribution of recurrent carries while preserving the architecture,
canonical flow path, and endpoint objective, the comparison isolates exposure to inference-induced
states as the intervention that unlocks recurrent test-time scaling.

\paragraph{Fixed-Point Forcing aligns convergence with correctness.}
Under FPF, additional recurrent updates reduce gold-target cross-entropy as solve rate increases,
directing convergent trajectories toward the correct solution rather than merely toward a
self-consistent prediction (Fig.~\ref{fig:fixed-point-stability-triptych}a,b). We measure convergence
by the token-averaged symmetric KL between adjacent predictive distributions,
$r_k=D_{\mathrm{SKL}}(p_k,p_{k-1})$. This residual changes from a chance-level correctness signal
under conventional self-conditioning to a nearly perfect one under FPF (AUROC $1.00$;
Fig.~\ref{fig:fixed-point-stability-triptych}c). Convergence thus becomes an observable signature of
successful reasoning, although not a formal certificate of correctness. Together, these findings
support the attractor picture in Fig.~\ref{fig:crown-jewel}: conventional self-conditioning admits
stable incorrect fixed points, whereas FPF reshapes the closed-loop dynamics so that correct
solutions behave as stable, error-correcting attractors. The depicted basins summarize this
empirical dynamical behavior rather than asserting a literal low-dimensional geometry.

\begin{figure*}[t]
  \centering
  \begin{minipage}[c]{0.38\textwidth}
    \centering
    (a) Distance to Ground Truth\par\vspace{0.25em}
    \includegraphics[width=\linewidth]{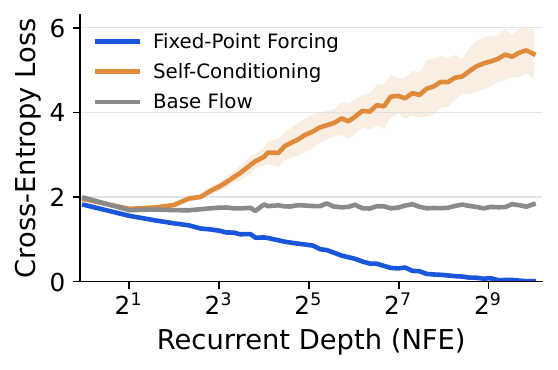}
  \end{minipage}\hfill
  \begin{minipage}[c]{0.38\textwidth}
    \centering
    (b) Solve Rate\par\vspace{0.25em}
    \includegraphics[width=\linewidth]{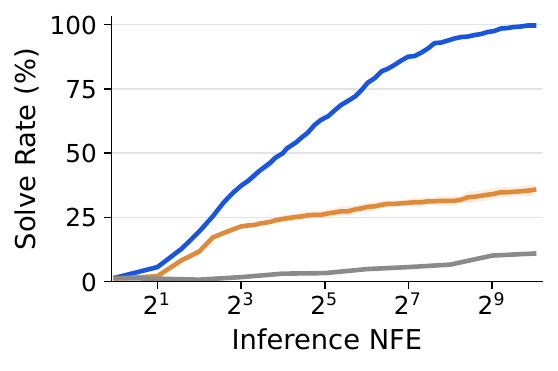}
  \end{minipage}\hfill
  \begin{minipage}[c]{0.20\textwidth}
    \centering
    (c)\par\vspace{0.6em}
    \scriptsize
    \setlength{\tabcolsep}{2pt}
    \begin{tabular}{lc}
      \toprule
      \textbf{Recipe} & \shortstack{\textbf{Residual}\\\textbf{AUROC}} \\
      \midrule
      Base Flow & $0.74$ \\
      Self-Cond. & $0.50$ \\
      \rowcolor{ourscol}\textbf{FPF} & $\mathbf{1.00}$ \\
      \bottomrule
    \end{tabular}
  \end{minipage}
  \caption{\textbf{Fixed-Point Forcing aligns recurrent convergence with correctness on Sudoku-Extreme.}
  (a) Under conventional self-conditioning, gold-target loss increases with depth as predictions
  stabilize at incorrect fixed points; FPF instead drives loss toward zero by making correct
  solutions attracting and spurious states unstable. (b) Consequently, FPF continues to improve
  solve rate with additional NFE, whereas self-conditioning saturates. (c) The residual between
  adjacent recurrent states becomes strongly predictive of correctness under FPF, indicating that
  the model converges primarily on correct solutions. Under conventional self-conditioning,
  convergence can instead occur at incorrect states.}
  \label{fig:fixed-point-stability-triptych}
\end{figure*}

\endgroup
\begin{table*}[t]
  \centering
  \small
  \setlength{\tabcolsep}{4pt}
  \resizebox{0.74\textwidth}{!}{%
    \begin{tabular}{lccc}
      \toprule
      & \textbf{Sudoku-Extreme} & \textbf{Zebra} & \textbf{Maze-Unique} \\
      \cmidrule(lr){2-2}\cmidrule(lr){3-3}\cmidrule(lr){4-4}
      \textbf{Model variant} & \textbf{Accuracy (\%)} & \textbf{Accuracy (\%)} & \textbf{Accuracy (\%)} \\
      \midrule
      Base Flow                           & $13.1 \pm 0.7$ & $62.3 \pm 29.1$ & $77.6 \pm 16.7$ \\
      + Self-conditioning                 & $32.6 \pm 3.7$ & $36.9 \pm 18.0$ & $96.2 \pm 3.0$ \\
      + Fixed-Point Forcing               & $\mathbf{99.2 \pm 0.3}$ & $\mathbf{99.9 \pm 0.2}$ & $\mathbf{98.0 \pm 1.8}$ \\
      \bottomrule
    \end{tabular}%
  }
  \caption{\textbf{Training-recipe ablation across three reasoning tasks.}
  Entries report the mean $\pm$ sample standard deviation over three training seeds. For each seed,
  we select the checkpoint and inference configuration with the highest validation solve rate.}
  \label{tab:component-ablation}
\end{table*}

\section{Related Work}
\label{sec:related}

\paragraph{Diffusion and Flow Language Models.}
A substantial body of work has developed diffusion-based alternatives to autoregressive modeling
for discrete data and language~\citep{austin2021structured,gat2024discreteflowmatching,
stark2024dirichletflowmatching}. Masked Diffusion Language Models generate sequences through
iterative denoising~\citep{sahoo2024simpleeffectivemaskeddiffusion}.
\citet{shi2025simplifiedgeneralizedmaskeddiffusion} derive a simplified continuous-time objective
as a weighted integral of cross-entropy losses and generalize the framework to state-dependent
masking schedules. Within this family,
\citet{kim2025trainworstplanbest} use adaptive token ordering to allocate computation, whereas
\citet{wang2025remaskingdiscretediffusionmodels} introduce remasking so earlier commitments can be
revised; Duo instead uses uniform categorical corruption, keeping every token revisable throughout
sampling~\citep{sahoo2025diffusion}. More recently, continuous flow language models have emerged as
competitive alternatives. FLM learns Gaussian-to-one-hot probability paths and FMLM distills them
into few-step maps~\citep{lee2026flowmaplanguagemodels}, while LangFlow and ELF define flows in
learned and contextual embedding spaces, respectively~\citep{chen2026langflowcontinuousdiffusionrivals,
hu2026elfembeddedlanguageflows}. FRMs are inspired by this diffusion-and-flow tradition, but use
the clean denoiser prediction as a directly decodable recurrent state and train the resulting
closed loop for iterative solution refinement rather than treating repeated evaluations only as
steps of a generative sampler.

\paragraph{Recurrent and Fixed-Point Reasoning.}
Recent structured reasoners scale test-time computation through weight-tied recurrence. HRM uses
hierarchical recurrent states~\citep{wang2025hierarchicalreasoningmodel}. TRM~\citep{
jolicoeurmartineau2025morerecursivereasoningtiny} and PTRM~\citep{
sghaier2026probabilistictinyrecursivemodel} repeatedly update compact answer and latent states.
EqR explicitly trains solution-aligned attractor dynamics~\citep{
huang2026equilibriumreasonerslearningattractors}. FPRM similarly observes that convergence becomes
predictive of correctness and uses this signal for adaptive halting~\citep{
movahedi2026fixedpointreasoners}.
FRMs share the emphasis on recurrent test-time scaling, attractor dynamics, and meaningful
convergence, but realize them within a self-conditioned flow language model: the flow denoiser's
clean prediction is both the recurrent input and an explicit solution candidate. Rather than
introducing a bespoke latent-state architecture, FRM retains the standard non-causal Transformer
and flow objective already demonstrated at language-model scale~\citep{
lee2026flowmaplanguagemodels,hu2026elfembeddedlanguageflows}, providing a natural path toward larger
models and less specialized domains. FPF further avoids backpropagation through the rollout:
model-generated states enter only through the conditioning channel, while every loss-bearing flow
state remains on the target-derived canonical path.

\paragraph{Learning from Model-Induced States.}
Conventional self-conditioning feeds a detached estimate of the clean sample into later denoising
steps, but trains that channel with a one-pass prediction rather than the recursively generated
predictions encountered during inference, as introduced for continuous encodings of discrete data
by Analog Bits~\citep{chen2023analogbitsgeneratingdiscrete}. This is an
instance of exposure bias: the model is evaluated on its own induced context without being trained
on that context~\citep{ross2011dagger,bengio2015scheduledsampling}. Concurrent work likewise frames
self-conditioned flow language models as fixed-point iterations for few-step generation
\citep{yoo2026selfconditionedflowmaplanguage}; we study structured reasoning and training on
model-induced recurrent states. Self Forcing closes the
analogous context mismatch in autoregressive video diffusion by training on self-generated rollout
context~\citep{huang2025selfforcingbridgingtraintest}. FPF can be understood as a self-forcing
variant specialized
to the recurrent conditioning channel of conditional discrete flows: it treats the carry, rather
than previous frames, as model-generated context. Unlike Self Forcing, FPF requires neither a
sequence-level distribution-matching objective nor differentiation through the rollout; it changes
only the carry distribution while preserving the canonical flow path and endpoint objective.

\section{Discussion}
\label{sec:discussion}
Flow Reasoning Models indicate that discrete flows can be a powerful and efficient model class for
structured reasoning. Across constraint satisfaction, relational deduction, and path finding,
FRMs achieve strong performance while retaining a favorable accuracy--compute tradeoff.
Conditioning discrete flows on their own past predictions creates an iterative refinement process,
while training on recursively generated states with Fixed-Point Forcing dramatically improves
performance and makes additional recurrent depth productive. Together, these results suggest that
recurrent reasoning need not rely on a specialized recurrent architecture: it can emerge from the
learned refinement dynamics of a general-purpose generative model. Scaling this approach to larger
models and less structured tasks is an important direction for future work.

\clearpage
\bibliography{references}

@misc{sahoo2024simpleeffectivemaskeddiffusion,
      title={Simple and Effective Masked Diffusion Language Models},
      author={Subham Sekhar Sahoo and Marianne Arriola and Yair Schiff and Aaron Gokaslan and Edgar Marroquin and Justin T Chiu and Alexander Rush and Volodymyr Kuleshov},
      year={2024},
      eprint={2406.07524},
      archivePrefix={arXiv},
      primaryClass={cs.CL},
      url={https://arxiv.org/abs/2406.07524},
}

@misc{shi2025simplifiedgeneralizedmaskeddiffusion,
      title={Simplified and Generalized Masked Diffusion for Discrete Data},
      author={Jiaxin Shi and Kehang Han and Zhe Wang and Arnaud Doucet and Michalis K. Titsias},
      year={2025},
      eprint={2406.04329},
      archivePrefix={arXiv},
      primaryClass={cs.LG},
      url={https://arxiv.org/abs/2406.04329},
}

@misc{wang2026remaskingdiscretediffusionmodels,
      title={Remasking Discrete Diffusion Models with Inference-Time Scaling},
      author={Guanghan Wang and Yair Schiff and Subham Sekhar Sahoo and Volodymyr Kuleshov},
      year={2026},
      eprint={2503.00307},
      archivePrefix={arXiv},
      primaryClass={cs.LG},
      url={https://arxiv.org/abs/2503.00307},
}

@misc{chen2023analogbitsgeneratingdiscrete,
      title={Analog Bits: Generating Discrete Data using Diffusion Models with Self-Conditioning},
      author={Ting Chen and Ruixiang Zhang and Geoffrey Hinton},
      year={2023},
      eprint={2208.04202},
      archivePrefix={arXiv},
      primaryClass={cs.CV},
      url={https://arxiv.org/abs/2208.04202},
}

@misc{hu2026elfembeddedlanguageflows,
      title={ELF: Embedded Language Flows},
      author={Keya Hu and Linlu Qiu and Yiyang Lu and Hanhong Zhao and Tianhong Li and Yoon Kim and Jacob Andreas and Kaiming He},
      year={2026},
      eprint={2605.10938},
      archivePrefix={arXiv},
      primaryClass={cs.CL},
      url={https://arxiv.org/abs/2605.10938},
}

@misc{chen2026langflowcontinuousdiffusionrivals,
      title={LangFlow: Continuous Diffusion Rivals Discrete in Language Modeling},
      author={Yuxin Chen and Chumeng Liang and Hangke Sui and Ruihan Guo and Chaoran Cheng and Jiaxuan You and Ge Liu},
      year={2026},
      eprint={2604.11748},
      archivePrefix={arXiv},
      primaryClass={cs.CL},
      url={https://arxiv.org/abs/2604.11748},
}

@misc{lee2026flowmaplanguagemodels,
      title={Flow Map Language Models: One-step Language Modeling via Continuous Denoising},
      author={Chanhyuk Lee and Jaehoon Yoo and Manan Agarwal and Sheel Shah and Jerry Huang and Aditi Raghunathan and Seunghoon Hong and Nicholas M. Boffi and Jinwoo Kim},
      year={2026},
      eprint={2602.16813},
      archivePrefix={arXiv},
      primaryClass={cs.CL},
      url={https://arxiv.org/abs/2602.16813},
}

@misc{wang2025hierarchicalreasoningmodel,
      title={Hierarchical Reasoning Model},
      author={Guan Wang and Jin Li and Yuhao Sun and Xing Chen and Changling Liu and Yue Wu and Meng Lu and Sen Song and Yasin Abbasi Yadkori},
      year={2025},
      eprint={2506.21734},
      archivePrefix={arXiv},
      primaryClass={cs.AI},
      url={https://arxiv.org/abs/2506.21734},
}

@misc{shah2024causallanguagemodeling,
      title={Causal Language Modeling Can Elicit Search and Reasoning Capabilities on Logic Puzzles},
      author={Kulin Shah and Nishanth Dikkala and Xin Wang and Rina Panigrahy},
      year={2024},
      eprint={2409.10502},
      archivePrefix={arXiv},
      primaryClass={cs.AI},
      url={https://arxiv.org/abs/2409.10502},
}

@misc{jolicoeurmartineau2025morerecursivereasoningtiny,
      title={Less is More: Recursive Reasoning with Tiny Networks},
      author={Alexia Jolicoeur-Martineau},
      year={2025},
      eprint={2510.04871},
      archivePrefix={arXiv},
      primaryClass={cs.LG},
      url={https://arxiv.org/abs/2510.04871},
}

@misc{huang2025selfforcingbridgingtraintest,
      title={Self Forcing: Bridging the Train-Test Gap in Autoregressive Video Diffusion},
      author={Xun Huang and Zhengqi Li and Guande He and Mingyuan Zhou and Eli Shechtman},
      year={2025},
      eprint={2506.08009},
      archivePrefix={arXiv},
      primaryClass={cs.CV},
      url={https://arxiv.org/abs/2506.08009},
}

@misc{huang2026equilibriumreasonerslearningattractors,
      title={Equilibrium Reasoners: Learning Attractors Enables Scalable Reasoning},
      author={Benhao Huang and Zhengyang Geng and Zico Kolter},
      year={2026},
      eprint={2605.21488},
      archivePrefix={arXiv},
      primaryClass={cs.LG},
      url={https://arxiv.org/abs/2605.21488},
}

@misc{kim2025trainworstplanbest,
      title={Train for the Worst, Plan for the Best: Understanding Token Ordering in Masked Diffusions},
      author={Jaeyeon Kim and Kulin Shah and Vasilis Kontonis and Sham Kakade and Sitan Chen},
      year={2025},
      eprint={2502.06768},
      archivePrefix={arXiv},
      primaryClass={cs.LG},
      url={https://arxiv.org/abs/2502.06768},
}

@misc{sghaier2026probabilistictinyrecursivemodel,
      title={Probabilistic Tiny Recursive Model},
      author={Amin Sghaier and Ali Parviz and Alexia Jolicoeur-Martineau},
      year={2026},
      eprint={2605.19943},
      archivePrefix={arXiv},
      primaryClass={cs.AI},
      url={https://arxiv.org/abs/2605.19943},
}

@article{sahoo2025diffusion,
      title={The Diffusion Duality},
      author={Subham Sekhar Sahoo and Justin Deschenaux and Aaron Gokaslan and Guanghan Wang and Justin Chiu and Volodymyr Kuleshov},
      journal={arXiv preprint arXiv:2506.10892},
      year={2025},
      url={https://arxiv.org/abs/2506.10892},
}

@inproceedings{lipman2023flowmatching,
  title     = {Flow Matching for Generative Modeling},
  author    = {Yaron Lipman and Ricky T. Q. Chen and Heli Ben-Hamu and Maximilian Nickel and Matt Le},
  booktitle = {International Conference on Learning Representations},
  year      = {2023},
  url       = {https://openreview.net/forum?id=PqvMRDCJT9t}
}

@article{albergo2025stochasticinterpolants,
  title   = {Stochastic Interpolants: A Unifying Framework for Flows and Diffusions},
  author  = {Michael S. Albergo and Nicholas M. Boffi and Eric Vanden-Eijnden},
  journal = {Journal of Machine Learning Research},
  volume  = {26},
  number  = {209},
  pages   = {1--80},
  year    = {2025},
  url     = {https://www.jmlr.org/papers/v26/23-1605.html}
}

@inproceedings{austin2021structured,
  title     = {Structured Denoising Diffusion Models in Discrete State-Spaces},
  author    = {Jacob Austin and Daniel D. Johnson and Jonathan Ho and Daniel Tarlow and Rianne van den Berg},
  booktitle = {Advances in Neural Information Processing Systems},
  volume    = {34},
  pages     = {17981--17993},
  year      = {2021},
  url       = {https://proceedings.neurips.cc/paper/2021/hash/958c530554f78bcd8e97125b70e6973d-Abstract.html}
}

@inproceedings{gat2024discreteflowmatching,
  title     = {Discrete Flow Matching},
  author    = {Itai Gat and Tal Remez and Neta Shaul and Felix Kreuk and Ricky T. Q. Chen and Gabriel Synnaeve and Yossi Adi and Yaron Lipman},
  booktitle = {Advances in Neural Information Processing Systems},
  volume    = {37},
  pages     = {133345--133385},
  year      = {2024},
  doi       = {10.52202/079017-4239},
  url       = {https://proceedings.neurips.cc/paper_files/paper/2024/hash/f0d629a734b56a642701bba7bc8bb3ed-Abstract-Conference.html}
}

@inproceedings{stark2024dirichletflowmatching,
  title     = {Dirichlet Flow Matching with Applications to {DNA} Sequence Design},
  author    = {Hannes Stark and Bowen Jing and Chenyu Wang and Gabriele Corso and Bonnie Berger and Regina Barzilay and Tommi Jaakkola},
  booktitle = {Proceedings of the 41st International Conference on Machine Learning},
  series    = {Proceedings of Machine Learning Research},
  volume    = {235},
  pages     = {46495--46513},
  publisher = {PMLR},
  year      = {2024},
  url       = {https://proceedings.mlr.press/v235/stark24b.html}
}

@inproceedings{wang2025remaskingdiscretediffusionmodels,
  title     = {Remasking Discrete Diffusion Models with Inference-Time Scaling},
  author    = {Guanghan Wang and Yair Schiff and Subham Sekhar Sahoo and Volodymyr Kuleshov},
  booktitle = {Advances in Neural Information Processing Systems},
  volume    = {38},
  pages     = {147282--147339},
  year      = {2025},
  doi       = {10.52202/085713-4927},
  url       = {https://proceedings.neurips.cc/paper_files/paper/2025/hash/d877ea0dd78bdbe54830670618c1de09-Abstract-Conference.html}
}

@misc{movahedi2026fixedpointreasoners,
  title         = {Fixed-Point Reasoners: Stable and Adaptive Deep Looped Transformers},
  author        = {Sajad Movahedi and Vera Milovanovi{\'c} and Shlomo Libo Feigin and Alexander Theus and Thomas Hofmann and Valentina Boeva and T. Konstantin Rusch and Antonio Orvieto},
  year          = {2026},
  eprint        = {2606.18206},
  archivePrefix = {arXiv},
  primaryClass  = {cs.AI},
  url           = {https://arxiv.org/abs/2606.18206}
}

@inproceedings{ross2011dagger,
  title     = {A Reduction of Imitation Learning and Structured Prediction to No-Regret Online Learning},
  author    = {St{\'e}phane Ross and Geoffrey Gordon and Drew Bagnell},
  booktitle = {Proceedings of the Fourteenth International Conference on Artificial Intelligence and Statistics},
  series    = {Proceedings of Machine Learning Research},
  volume    = {15},
  pages     = {627--635},
  publisher = {PMLR},
  year      = {2011},
  url       = {https://proceedings.mlr.press/v15/ross11a.html}
}

@inproceedings{bengio2015scheduledsampling,
  title     = {Scheduled Sampling for Sequence Prediction with Recurrent Neural Networks},
  author    = {Samy Bengio and Oriol Vinyals and Navdeep Jaitly and Noam Shazeer},
  booktitle = {Advances in Neural Information Processing Systems},
  volume    = {28},
  year      = {2015},
  url       = {https://proceedings.neurips.cc/paper/2015/hash/e995f98d56967d946471af29d7bf99f1-Abstract.html}
}

@misc{yoo2026selfconditionedflowmaplanguage,
      title={Self-conditioned Flow Map Language Models via Fixed-point Flows},
      author={Jaehoon Yoo and Wonjung Kim and Floor Eijkelboom and Chanhyuk Lee and Nicholas M. Boffi and Seunghoon Hong and Jinwoo Kim},
      year={2026},
      eprint={2607.00714},
      archivePrefix={arXiv},
      primaryClass={cs.CL},
      url={https://arxiv.org/abs/2607.00714},
}
\bibliographystyle{template/iclr2026_conference}

\appendix

\section{Experimental Details}
\label{app:experiments}

\subsection{Datasets}
\label{app:datasets}

\paragraph{Sudoku-Extreme.}
We use Sudoku-Extreme to test long-horizon constraint satisfaction on hard $9\times9$ puzzles that
require extended chains of deduction and search. The benchmark was introduced with the Hierarchical
Reasoning Model as a small-data test of recurrent reasoning
\citep{wang2025hierarchicalreasoningmodel}; we represent each released puzzle as its unique solution
together with a mask that clamps the given clues. For each training seed, we draw a
difficulty-balanced subset of 1,000 puzzles from the official training split and apply random Sudoku
symmetries as augmentation. We select checkpoints on a fixed held-out validation subset and report
exact-grid solve rate on a fixed 1,000-puzzle subset of the official test split.

\paragraph{Zebra.}
We use Zebra to test whether recurrent refinement transfers from grid-local constraints to
heterogeneous relational reasoning. Introduced as a reasoning benchmark by
\citet{shah2024causallanguagemodeling}, the dataset contains approximately 1.5 million training and
0.1 million test instances spanning three to six houses and attributes. Following the uniform
$3\times3$--$6\times6$ mixture introduced by \citet{shah2024causallanguagemodeling} and used by
\citet{kim2025trainworstplanbest}, we train and evaluate across all four puzzle sizes rather than
restricting the benchmark to a fixed $5\times5$ slice. We retain each puzzle's original clue
serialization but replace its variable, solver-ordered trace with a canonical attribute-by-house
solution grid. We report exact-grid solve rate on the mixed held-out split.

\paragraph{Maze-Unique.}
We use Maze-Unique to test structured path finding under an unambiguous exact-solve criterion.
Released with Equilibrium Reasoners \citep{huang2026equilibriumreasonerslearningattractors}, this
$30\times30$ path-finding benchmark provides official, revision-pinned splits of 1,000 training and
1,000 test mazes; we encode walls, start, and goal as clamped cells and generate the remainder of the
grid. We prefer it to the more common Maze-Hard benchmark because every instance has a unique
solution path, making exact-grid accuracy coincide with path validity and optimality instead of
penalizing an alternative valid route. It also permits direct comparison with recent recurrent
reasoners evaluated on the same split.

\subsection{Models and Training}
\label{app:implementation}
All experiments use a non-causal DiT backbone and the categorical clean-prediction objective.
AdamW with $\beta=(0.9,0.999)$, gradient clipping at $1.0$, EMA decay $0.9999$, dropout $0.1$, and a
constant learning-rate schedule are shared across datasets. The learning rate, weight decay, batch
size, warmup, training budget, and architecture vary by dataset and are reported in
Table~\ref{tab:training-hyperparameters}. In every case, Stage~A trains self-conditioning from
scratch; Stage~B initializes from the selected Stage-A
weights with a fresh optimizer and EMA state before applying FPF.
For each training seed, we select the checkpoint and inference configuration (including sampler and
NFE) with the highest validation solve rate. Table~\ref{tab:component-ablation} aggregates the
resulting selected operating points across seeds.

\begin{table*}[t]
  \centering
  \small
  \begin{tabular*}{\textwidth}{@{\extracolsep{\fill}}lccc@{}}
    \toprule
    \textbf{Setting} & \textbf{Sudoku-Extreme} & \textbf{Zebra $3\times3$--$6\times6$} & \textbf{Maze-Unique} \\
    \midrule
    \multicolumn{4}{l}{\textit{Architecture}} \\
    Parameters & $7.0$M & $25.0$M & $8.0$M \\
    Hidden size / blocks / heads & $256$ / $7$ / $8$ & $384$ / $12$ / $12$ & $256$ / $8$ / $8$ \\
    Sequence length / vocabulary & $81$ / $10$ & size-dependent / $21$ & $900$ / $6$ \\
    \midrule
    \multicolumn{4}{l}{\textit{Optimization}} \\
    Optimizer / betas & AdamW / $(0.9,0.999)$ & AdamW / $(0.9,0.999)$ & AdamW / $(0.9,0.999)$ \\
    Learning rate / weight decay & $3\times10^{-4}$ / $0$ & $3\times10^{-4}$ / $0$ & $1\times10^{-4}$ / $1.0$ \\
    Global batch size per step & $128$ & $128$ & $64$ \\
    Warmup steps & $0$ & $0$ & $500$ \\
    Gradient clip & $1.0$ & $1.0$ & $1.0$ \\
    EMA decay & $0.9999$ & $0.9999$ & $0.9999$ \\
    Dropout & $0.1$ & $0.1$ & $0.1$ \\
    Learning-rate schedule & constant & constant & constant \\
    Stage-A / Stage-B iterations & $100$K / $100$K & $200$K / $200$K & $300$K / $100$K \\
    \midrule
    \multicolumn{4}{l}{\textit{Self-conditioning and FPF}} \\
    Self-conditioning probability & $0.5$ & $0.5$ & $0.5$ \\
    FPF rollout probability / depth & $0.5$ / $16$ & $0.5$ / $16$ & $0.5$ / $16$ \\
    Supervised states / noise scale & $4$ / $0$ & $4$ / $0$ & $4$ / $0$ \\
    \bottomrule
  \end{tabular*}
  \caption{\textbf{Per-dataset architecture and training hyperparameters.} Global batch size is the
  number of puzzles used in each optimizer step.}
  \label{tab:training-hyperparameters}
\end{table*}

\subsection{Baseline Provenance and Comparability}
\label{app:baseline-accounting}
We organize baselines into three families because they make different architectural commitments.
The autoregressive baseline is a plain causal decoder-only transformer trained with
teacher-forced next-token cross-entropy over a clue-grid-then-solution-grid sequence: no denoising,
no recurrence, and no mechanism to revise an already-committed token, making it the naive baseline
the rest of the comparison is meant to contextualize~\citep{shah2024causallanguagemodeling}. Diffusion
language models and flows use a general-purpose generative denoising formulation that can share the
same non-causal DiT backbone as FRM. Small recurrent reasoning models instead introduce
task-oriented recurrent state, update, or halting mechanisms. This
grouping clarifies the role of each comparison; it does not by itself imply matched capacity or
compute, so Table~\ref{tab:best-performance-compute} reports the model size used on each benchmark and
Appendix~\ref{app:compute-accounting} reports inference cost.

\paragraph{Accuracy provenance.}
Unless noted here, entries in Table~\ref{tab:best-performance-compute} are measurements from our
evaluation harness using checkpoints that we trained or official checkpoints that we evaluated.
For Zebra, the FMLM, FLM, and FRM entries are our measurements. The daggered MDLM and adaptive-MDLM
accuracies are published values from \citet{kim2025trainworstplanbest} on Shah's uniform
$3\times3$--$6\times6$ mixture; we could not reproduce those reported accuracies in our pipeline.
Our weight-backed Zebra MDLM reproduction peaks at $49.2\%$ under our shorter training budget and
supplies the measured operating points used for efficiency analysis. We omit ReMDM and specialized
recurrent reasoners on Zebra because we did not obtain a reliable result under the same mixed
protocol; the dashes therefore indicate unavailable comparisons, not zero accuracy. For
Maze-Unique, EqR is a published reference value, whereas TRM is our $0.264$M-parameter
reproduction trained and evaluated on the Maze-Unique split, reaching $77.9\%$ exact solve rate.
The $44.9\%$ result reported by the original TRM work is for Maze-Hard, not Maze-Unique, and is
therefore excluded from this table. The remaining Maze-Unique entries are measured from our trained
reproductions or evaluated checkpoints. Thus,
published accuracies support the headline comparison, whereas every point in an accuracy--compute
frontier is produced from a checkpoint available to our profiler.

\subsection{Inference Compute Accounting}
\label{app:compute-accounting}

\paragraph{Number of function evaluations.}
We report the realized number of learned-model evaluations required to solve one puzzle. For our
Euler, SDE, and held-time samplers, NFE counts every iterative update and the final readout
evaluation; for methods with adaptive stopping, branching, or breadth, we count all evaluations
actually executed for that instance. NFE exposes the number of sequential model calls and is easy
to reproduce, but it is not a sufficient cross-method compute measure: one evaluation can differ
substantially in cost across architectures, sequence lengths, model sizes, and internally recurrent
networks. The appendix table reports \emph{solved-puzzle mean NFE}: for each puzzle that is solved at
some point along its evaluated trajectory, we record the first NFE at which it is solved and average
over solved puzzles only. This is an oracle trajectory diagnostic unless correctness can be detected
without access to the ground truth; it should therefore not be interpreted as deployable early
stopping for every method.

\paragraph{Floating-point operations.}
We therefore use total FLOPs per puzzle as the primary hardware-independent compute measure. We
profile an actual batch-one forward pass with PyTorch's operator-level FLOP counter, disabling fused
attention so that its matrix multiplications are visible, and multiply the measured per-evaluation
cost by the realized NFE; methods with nonuniform calls are profiled and summed by call type. This
quantity should be interpreted as a lower bound on the work required to solve an instance: it counts
floating-point arithmetic but not memory traffic, kernel-launch and framework overhead, control
flow, communication, or data movement. We therefore present the full accuracy--compute frontiers
rather than compressing each method to a single operating-point ratio.

\section{Additional Accuracy and Ablation Results}
\label{app:headline-support}
\label{app:component-support}
This section documents the headline accuracy--compute comparison and training-recipe ablation;
Appendix~\ref{app:experiments} gives datasets, checkpoint provenance, and FLOP accounting.

\subsection{Accuracy--Compute Evaluation}
Each frontier contains measured operating points only. We vary each checkpoint's native compute
control and charge every learned-model call. Fixed-depth methods use the requested depth; adaptive
methods use realized mean computation. We neither extrapolate unsupported method--dataset pairs nor
include published accuracies lacking a compatible profiled checkpoint, although those values may
still appear in Table~\ref{tab:best-performance-compute}.

\begin{table}[H]
  \centering
  \small
  \setlength{\tabcolsep}{5pt}
  \begin{tabular*}{0.94\textwidth}{@{\extracolsep{\fill}}lll@{}}
    \toprule
    \textbf{Method} & \textbf{Inference-time control} & \textbf{Charged computation} \\
    \midrule
    FMLM / FLM & Flow-integration steps & All denoiser evaluations \\
    MDLM & Denoising steps & All denoiser evaluations \\
    ReMDM & Denoising and remasking steps & All denoiser and remasking evaluations \\
    Adaptive MDLM & Update cap and confidence rule & Realized adaptive evaluations \\
    HRM / TRM / FPRM & Recurrent depth & All recurrent model evaluations \\
    EqR & Equilibrium depth and breadth & All iterations and evaluated branches \\
    FRM (ours) & Recurrent depth and sampler & All recurrent denoiser evaluations \\
    \bottomrule
  \end{tabular*}
  \vspace{0.5em}
  \caption{\textbf{Inference controls used to construct accuracy--compute frontiers.} Every
  learned-model invocation is included in the reported computation. Exact evaluated grids are
  checkpoint- and dataset-specific.}
  \label{tab:frontier-inference-controls}
\end{table}

\paragraph{Adaptive computation.}
\label{app:mdm-floor}
Token-adaptive methods can incur a nonzero minimum cost even at their lowest nominal setting. We
therefore plot measured realized FLOPs rather than the maximum-step label.

\subsection{Training-Recipe Results}
Table~\ref{tab:component-ablation} reports three-seed means and standard deviations for every
dataset. We report aggregate statistics consistently and omit individual-seed values.

\section{Additional Dynamics Analysis}
\label{app:mechanism-support}
We distinguish productive refinement from convergence to an incorrect fixed point using the
following depth-dependent diagnostics.

\paragraph{Depth-dependent performance and distance to the solution.}
At each depth, we record exact solve rate and token-averaged cross-entropy to the one-hot solution
on the same pinned examples. Thus changes isolate additional recurrent computation. In
Fig.~\ref{fig:fixed-point-stability-triptych}, FPF improves accuracy while driving ground-truth
loss downward; conventional self-conditioning becomes increasingly confident in incorrect states.

\paragraph{Convergence residual and correctness prediction.}
For adjacent distributions $p_k$ and $p_{k-1}$, we measure the token-averaged symmetric-KL
residual $r_k=D_{\mathrm{SKL}}(p_k,p_{k-1})$ and pair it with the decoded state's exact-solve label.
A near-zero residual indicates convergence, not necessarily correctness. Conventional
self-conditioning yields chance-level discrimination (AUROC $0.50$); under FPF, convergence is
strongly predictive of correctness (AUROC $1.00$) on Sudoku-Extreme.

\paragraph{Interpretation.}
Exact solve rate remains primary; loss-to-gold and adjacent-state residuals reveal whether depth
corrects the solution or settles into a confidently incorrect fixed point.
\end{document}